\documentstyle[11pt,epsfig]{article}
\newcommand{\be}{\begin{equation}}
\newcommand{\ee}{\end{equation}} 
\newcommand{\eei}{\end{equation}\indent\indent}
\newcommand{\bc}{\begin{center}}
\newcommand{\ec}{\end{center}}
\newcommand{\ber}{\begin{eqnarray}}
\newcommand{\ear}{\end{eqnarray}}
\newcommand{\ba}{\begin{array}}
\newcommand{\ea}{\end{array}}

\def\case#1/#2{\textstyle\frac{#1}{#2} }
\begin{document}
\title{P-model Alternative to the T-model}
\author{Mark D. Roberts, \\
117 Queens' Road, Wimbledon, London SW19 8NS\\
Email mdr@ic.ac.uk,  http://cosmology.mth.uct.ac.za/$\sim$ roberts}
\maketitle
\bc Eprints:     http://arXiv.org/abs/cs.CL/9811018  
                 http://cogprints.soton.ac.uk/abs/cog00000933 \ec
    Comments:  28 pages  73262 bytes,  six diagrams,  53 references,\\
some small changes from the previous version, 
background to this work is described:
http://cosmology.mth.uct.ac.za/~roberts/pastresearch/pmodel.html
\bc Mathematical Reviews Subject Classification: http://www.ams.org/msc/ 
     03D55,  92K20,  68S05.\ec 
\bc ACM Classification:  http://www.acm.org/class/1998/overview.html
    I.2.7,J.4,I.2.6\ec
\bc Keywords:  T-model,Movement,  Frege Representation.\ec
\begin{abstract}
Standard linguistic analysis of syntax uses the T-model.   This model 
requires the ordering:  D-structure $>$ S-structure $>$ LF,   
where D-structure is the sentences deep structure,  
S-structure is its surface structure,  and LF is its logical form.   
Between each of these representations there is movement which alters 
the order of the constituent words;  movement is achieved using the principles
and parameters of syntactic theory.   Psychological analysis of sentence 
production is usually either serial or connectionist.   Psychological serial
models do not accommodate the T-model immediately so that here a new model 
called the P-model is introduced.   The P-model is different from previous 
linguistic and psychological models.   Here it is argued that the LF 
representation should be replaced by a variant 
of Frege's three qualities (sense,  reference,  and force),  
called the Frege representation or F-representation.
In the F-representation the order of elements is not necessarily the same as 
that in LF and it is suggested that the correct ordering is: 
F-representation $>$ D-structure $>$ S-structure.   
This ordering appears to lead to a more natural 
view of sentence production and processing.   Within this framework movement 
originates as the outcome of emphasis applied to the sentence.   The 
requirement that the F-representation precedes the D-structure needs a picture
of the particular principles and parameters which pertain to movement of words
between representations.   In general this would imply that there is a 
preferred or optimal ordering of the symbolic string in the F-representation.
The standard ordering is retained because the general way of producing 
such an optimal ordering is unclear.   In this case it is possible to produce 
an analysis of movement between LF and D-structure similar to the usual 
analysis of movement between S-structure and LF.   
The necessity of analyzing corrupted data
suggests that a maximal amount of information about 
a language's grammar and lexicon is stored.    
\end{abstract}
{\small\tableofcontents}
\section{Introduction.}
\label{sec:intro}
\subsection{Forward.}
For the purposes of theory language can be split up into segments:
paragraphs, sentences and so forth.   It is not always clear what the segments
are in natural language,  especially spontaneous speech,
see for example Rischel (1992) \cite{rischel}.                    
Linguists and psycholinguists give priority to the
analysis of sentences and approach this in different ways,  for example,  
there are a variety of ways of approaching sentence word order.
Linguists usually approach word order by invoking the T-model 
in which various principles are used to change word order from 
a primitive form (D-structure) to the audible or written form (S-structure);  
linguists do not seem to realize that the T-model is a psycholinguistic 
processing model.   Other ways of accounting for word order include 
the Markov cascade models of Brants (1999) \cite{brants}.

Linguistics is split up into several subdisciplines which include:
phonology,  morphology,  social,  historical,  semantics,  and syntax.
According to James McCloskey (1988) \cite{mccloskey}:
\begin{quote}
the study of syntax has always been a more acrimonious business.....
than the pursuit of [its] sister disciplines in linguistics.
\end{quote}
Modern syntax grew from the need to record the grammars of North American 
Indians whose languages where rapidly becoming extinct,
much of this work was done by Bloomfield in the 1920's.
In the 1950's Chomsky started applying an analogy with pure mathematical
category theory to syntax - then syntax began to take its modern
form.  The way that syntactical investigations usually take place is by 
analyzing contrived sentences,  rather than naturally occurring sentences.
Similarly psychology is split up into several subdisciplines which include:
social,  developmental,  psychometric,  neuropsychology,  and cognitive.

Linguistics and psychology are not as sociologically closely related 
as one might expect.    For example,  psycholinguists rarely
produce phrase trees of their test sentences:  this would be the starting 
point of any syntactic analysis;  an exception to ignoring the work of
linguists being Hall (1995) \cite{hall} who discusses potential contributions 
of psycholinguistic techniques to Universal Grammar.
Similarly syntactians rarely refer to the 
measurements of psycholinguists.   Some well known linguistic textbooks 
on syntax,  for example Chomsky (1986) \cite{bi:chomsky},  
invoke and usually start with a processing model called the T-model.   
Here I point out that this is indeed a psycholinguistic model,  
and can be subject to the methodology of that discipline.  
This model requires 
the ordering: $D(eep)-Structure>S(urface)-Structure>LF(logical form)$.   
Between each of these representations there is movement,  described by 
various principles,  which alters the order between constituent words.
That grammar should relfect more closely the workings of the human
parser has been suggested by Phillips (2001) \cite{phillips} and
Richards (1999) \cite{richards}.

Frege analyzed the meaning of a sentence to 
depend on three qualities:  sense,  reference,  and force.   
In \S2.2 it is argued that Frege's {\em three} qualities which describe 
a sentence:  sense,  reference,  and force,  should be replaced by the
{\em five} qualities:  external referents,  lexical referents,  formal
declarants,  formal string,  and force.   
These qualities form a 
representation here called the F(rege)-representation.
In the F-representation the order of elements is not necessarily the same as 
in the LF.  It is then suggested that the correct ordering is:  
$F-representation>D-structure>S-structure$.   This ordering leads to a more 
natural view of sentence production (or processing) called the P-model.   
Within this framework movement originates as the outcome of the 
emphasis applied to the sentence;  rather than as it occurs in linguistic 
models where movement is unmotivated and {\it ad hoc}.
The requirement that the F-representation 
precedes the D-structure needs a picture of the principles and parameters 
which pertain to movement of words between representations.   In general this 
would imply that there is a preferred or optimal ordering of the symbol string
in the F-representation. The general way of producing such an optimal ordering
is unclear;  but might be found by invoking an extremal principle as 
discussed in \S3.1.   In \S6 a new model called the P-model is presented,
this model uses the standard ordering,  as the optimal ordering is still
unknown.   For the P-model it is
possible to produce an analysis of movement between LF and D-structure similar
to the usual analysis of the movement between S-structure and LF.

In \S3.3 it is suggested that a maximal amount 
of information about a language's grammar and lexicon (vocabulary) are stored.
At first sight this might seem inefficient,  
but could occur because it allows 
for a quick analysis of speech which is often only partial heard.
\subsection{Language Acquisition.}
McDonald (1997) \cite{mcdonald} reviews how language learners master the formal
structure of their language.   She investigates {\em Three} possible routes 
to the acquisition of linguistic structure:   
{\em firstly} the use of prosodic and phonological information,   
{\em secondly} the use of function
words to syntactically classify co-occurring words and phrases,  and 
{\em thirdly}  the use of morphology internal to the lexical items to determine
language structure,  and the productive recombination of these subunits 
in new items.   Evidence supporting these three routes comes from normal 
language acquirer's and from several special populations,  including learners 
given improvised output,   learners with Downs syndrome,  and late 
learners of first and second languages.   Further evidence for the three 
routes comes from artificial language acquisition experiments and computer 
simulations,  see also Ferro {\it et al} \cite{FVY}.   
Language acquisition has also been reviewed by 
Gleitman and Bloom (1998) \cite{GB}.
There is also the problem of how the segments of language,  
as discussed in \S5.1, occur in language acquisition. 
Josephson and Blair (1998) \cite{JB} view language acquisition primarily as
an attempt to create processes that connect together fruitfully linguistic
input and other activity.   From a philosophers point of view this is a 
{\sc coherence} theory of language,   see the discussion in 
Roberts (1998) \cite{rad}\S6.1.
Hymans (1985) \cite{hymans} discusses how parameters 
are set in language acquisition.
Valian (1990) \cite{valian} argues that in a child's acquisition of whether
to have a null subject or not,  
there is an initial dual switch allows both options;
the previously accepted model being either null subjects or not at one
value being predetermined at one choice.
There is also the question of how language evolved in the first place,
the main differences being over whether most of it evolved recently {\em or}
whether it evolved gradually,  
see the review of MacWinney (1998) \cite{macwinney}.
\subsection{The T-Model.}
The T-model is the basic framework within which linguistic syntax is currently
understood.   There are various levels with different word order,  the word 
order being altered by movement.   The object of linguistic syntax is to
find the rules which describe movement.   The T-model is usually pictured by 
diagram one of an upside down Y,  
(not an upside down T from which its name derives)
see for example:  
Chomsky (1986) \cite{bi:chomsky} p.68,  
Cook (1988) \cite{bi:cook} p.31,  
Haegeman (1994) \cite{bi:haegeman} p.493,
Hornstein (1995) \cite{bi:hornstein} p.2.
\begin{figure}
   \centering
   {\epsfig{figure=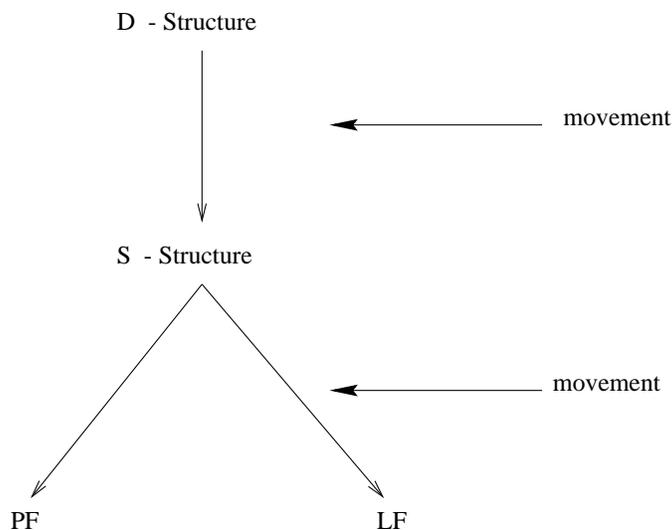,width=.7\textwidth}}
   \caption{Diagram One - The T-model.}
\end{figure}
Brody (1998) \cite{brody} presents a system of principles relating the LF
representation to lexical items that are compatible to his assumption of 
no externally forced imperfections in syntax;  his assumption is a 
generalization of the linguistic projection principle.

The T-model illustrated in diagram one consists of several levels.
The D-structure (deep structure) level is supposed to hold a given sentence in
a primitive form.   The D-structure level cannot be the same as the message 
level in serial psychological models as there individual words 
are supposed to be already delineated.   
In the Garret model (1980) \cite{bi:garret}
(see also Garman (1990) \cite{bi:garman} p.394) the D-structure 
level could be interpreted as being about midway in the sentence level.   
This level is then subject to various rules pertaining to how the order of 
the words can be moved.   These rules make up the bulk of the principles and 
parameters approach to linguistic syntactical analysis.    The communicative
purpose of such movement is to alter the emphasis of D-structure.   The
consequence of movement on the D-structure level is to produce another level
called the S-structure (surface structure) level.   The S-structure sentences 
are physically realized by the PF (phonetic form) and this corresponds to the 
positional level representation in serial models.   
To produce a {\it post-hoc} 
analysis of S-structure for its formal content it is postulated,  by analogy 
with movement from S to D-structure,  that movement again occurs to bring the 
sentence into its LF (logical form).   The "T" diagram is sometimes made more 
complex by the addition of other factors,  
e.g. Cook(1988) \cite{bi:cook} p.33.
Here movement from S-structure to LF is illustrated with examples
of quantifier-raising and Wh-raising taken from 
Haegeman (1994) \cite{bi:haegeman} Ch.9.
\subsection{Quantifier-Raising.}
Consider the surface structure sentence
(compare Haegeman (1994) \cite{bi:haegeman} p.489 eq.3):
\be
Jones~ saw~ everyone.
\label{eq:jse}
\ee
To this surface structure word order movement is applied to yield the LF
word order given by the string \ref{eq:qrlf}.
This can be represented by diagram two
\begin{figure}
   \centering
   {\epsfig{figure=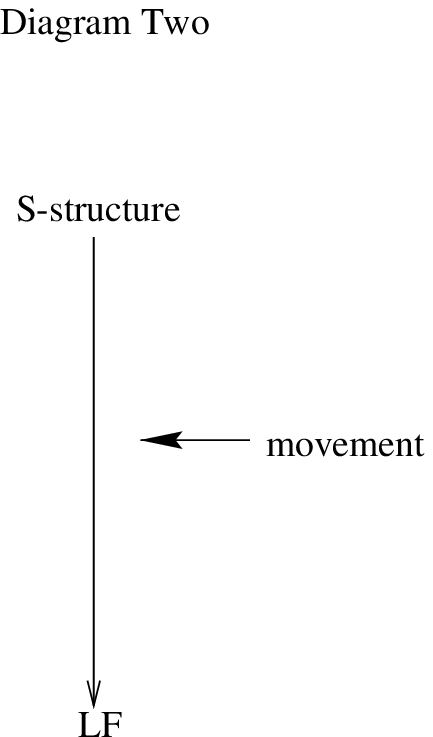,height=0.5\textwidth}}
   \caption{Diagram Two - Quantifier Raising}
\end{figure}
which can be expressed in terms of symbols
\be
\forall x ,( x\varepsilon H \rightarrow J \varepsilon Sx),
\label{eq:faf}
\ee
where H,J and S denote ``human'',  ``Jones'',  and ``saw'' respectively,
and $\forall$ denotes 'for all'.
Using traces this can be expressed in terms of words
\be
For~ all~ x~ it~ is~ the~ case~ that~ if~ x~ is~ human~ then~ Jones~ saw~ x.
\ee
or in terms of the quantifier $everyone_{i}$   
\be
[Everyone_{i}[Jones~saw~ x_{i}]].
\label{eq:qrlf}
\ee
The idea here is that the universal quantifier $everone_{i}$
can be put first in the LF representation.    
The universal quantifier could also be put last,  
but by convention it is taken to come first.   
To achieve this there is movement from $everyone_{i}$ being last 
to being first in the S-structure this leaves a trace $x_{i}$
in place of $everyone_{i}$
\subsection{Wh-Raising.}
Consider the sample surface structure question
(compare Haegeman (1994) \cite{bi:haegeman} p.494 eq.9):
\be
Who~ did~ Jones~ see?
\ee
In this case the word order of the surface structure 
and the LF remain the same.   In term of symbols
\be
Wh (x)~~~,~~~x\varepsilon H~~~, J \varepsilon Sx
\ee 
where $Wh(x)$ denotes ``Who \ldots ''.
Using traces this can be expressed in terms of words
\be
For~ which~ x, ~~ x~ is~ human, ~~ is~ it~ the~ case~ that~ Jones~ saw~ x?
\ee
This has S-structure representation
\be
[_{CP}Who_{i}~did [_{IP} Jones~ see~ t_{i}]]?
\ee
and LF representation
\be
[_{CP}Who_{i}~ did [_{IP} Jones~ see~ x_{i}]]?
\ee
The S-structure and the LF have the same form,  but with different traces.
The S-structure ordering of Wh-phrases is not the same in all languages;
in some Wh-words do not appear on the left,  so that in these cases both
the S-structure and the LF would not have the same form:  having LF depending
on specific languages is contrary to its name.   There can be
ambiguity in the scope (or domain of applicability) of the quantifiers;
this is illustrated by the sentence 
(compare Haegeman (1994) \cite{bi:haegeman} p.490 eqs.4 and 5)
\be
Everyone~ saw~ someone.                               
\label{eq:ess}
\ee
which has the two interpretations:
\be
For~ every~ x~ there~ is~ some~ y~ such~ that~ 
it~ is~ the~ case~ that~ x~  saw~ y.
\label{eq:fex}
\ee
and 
\be
There~ is~ some~ y,~~  such~ that~ for~ every~ x, 
~~ it~ is~ the~ case~ that~ x~ saw~ y.
\label{eq:fig}
\ee
\section{F-level Generalization of Logical Form.}    
\subsection{Drawbacks with the Standard Approach\\ to Logical Form.}
The notion of logical form is described in May (1985) \cite{bi:may} and  
Hornstein (1995) \cite{bi:hornstein},
Stanley (1998) \cite{stanley} discusses the origin of logical form.
Logical form as currently understood has at least two drawbacks.   
The {\it first} is that it is restricted - in the sense that it is based on 
the simple calculi of logic such as the predicate and propositional calculi. 
For example sentences such as
\be
The~ probability~ of~ snow~ is~ 80\%.                     
\label{eq:prob}
\ee
require an understanding of probability and hence of real numbers, $\cal{R}$,
which have continuous properties as opposed to the discrete properties of
simple calculi and standard LF (logical form).   
This implies that sentence \ref{eq:prob} 
requires a larger formal structure,  encompassing real numbers $\cal{R}$, 
than is usually included in LF.    
Sentence \ref{eq:prob} can be represented
\be
Prob,~~~   p\varepsilon [0,1]    ,    ~~~~~~~~(snow=0.8)              
\ee
with similar representations for other formal mathematical statements such as
occur in fuzzy logic.   The {\it second} drawback is that a string such as
\be
(x\varepsilon H\rightarrow J\varepsilon Sx)    \forall x,
\label{eq:fal}
\ee
unambiguously means the same as equation \ref{eq:faf};  
but the symbol order is different,
$\forall x$ (for all x) coming first in equation \ref{eq:faf} 
and last in equation \ref{eq:fal}.
In the principles and parameters approach the order of the symbols is 
essential,  otherwise it is hard to know where to insert traces,
compare \S1.4\&1.5.
\subsection{A New Approach to Logical Form.}
To overcome these drawbacks consider Frege's approach to the meaning of a
sentence.   To quote Dummet (1973) \cite{bi:dummet} p.83
\begin{quote}
Frege drew,  within the intuitive notion of meaning,  a distinction 
between three ingredients:  sense,  tone and force.
\end{quote}
Here the variation of these that is used can be represented by 
diagram three:
\begin{figure}
   \centering
   {\epsfig{figure=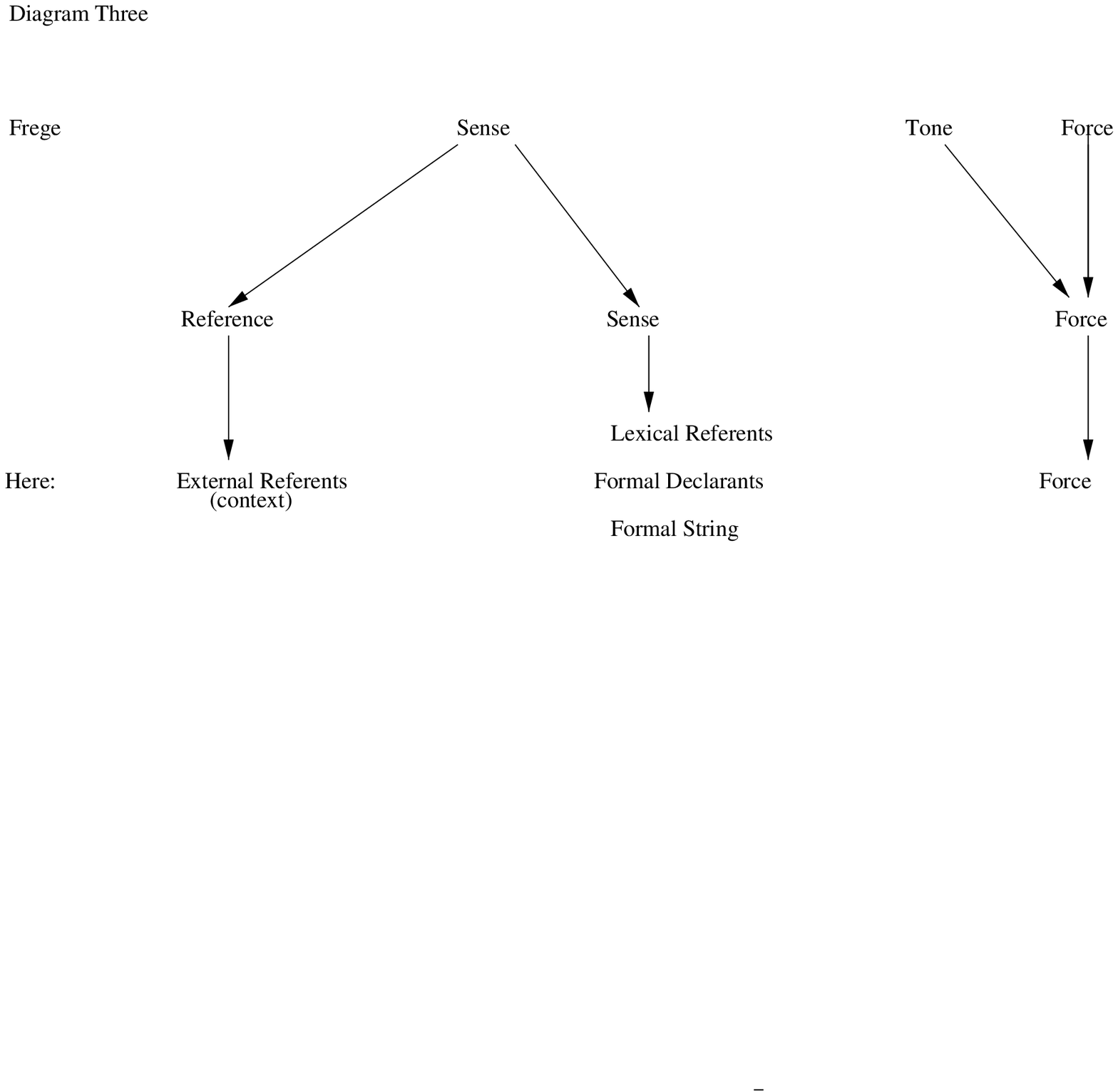,width=1.0\textwidth,}}
   \caption{Diagram Three - modification of Frege's sentence content
 from sense,  tone \& reference to five factors.}
\end{figure}
Reference and force are essentially unchanged,  but sense is decomposed into 
three.   Lexical referents are words and the concomitant symbol used in the 
formal language.   Formal declarants are similar to the beginning of computer
programmes where 
the following three things are specified
the language used,  the parameters used,  and the scope 
(local or global,  or other specified ordering to give precedence in case of
ambiguity)  are specified.   The formal string is an ordered set of symbols
which are well formed (well defined) in the language.  For example the sense
for equation \ref{eq:jse} would be decomposed as follows
\newline
\ber
\begin{array}{lll}
Lexical Referents  &:&     g = Jones,~   S = saw,~  h = human,\\
Formal Declarants  &:&     Predicate~ Calculus,~    x \varepsilon H,\\
Formal String      &:&    
\forall x(x\varepsilon H\rightarrow J\varepsilon Sx)                    .
\end{array}
\ear
Another advantage of the new approach is that sentence
\ref{eq:ess} is ambiguous leading to either 
\ref{eq:fex} or \ref{eq:fig}.   This ambiguity can be 
removed by limiting the scope of the variables in the formal declarants.
\section{Maximal Verses Minimal Encoding of Information.}
\subsection{Extremal Principles.}
In physics there are {\it least action principles} which,  when the action is
minimal,  give differential equations which describe the dynamics of systems.
The analogy has been carried through to other areas of science,
see for example Roberts (1998) \cite{ultra}\S3\&\S1\P3 and references therein.
There is a minimalist program in theoretical linguistics,  
Lasnik (1998) \cite{lasnik} and Culicover (1998) \cite{culicover},  
which invokes an {\em economy principle} where the steps,  symbols  
and representations in the principle and parameters approach are minimal.
Gibson (1998) \cite{gibson} proposes a theory 
which invokes {\em economy of processing};  this theory relates 
sentence processing to available computational resources.   
The computational resources have two components,  
{\it firstly} an {\sc integration cost component} 
and {\it secondly} a
{\sc memory cost component} which are quantified in the number of 
syntactic categories that are necessary to complete the current input string
as a grammatical sentence.   These cost components are influenced by
{\em locality} which entails both 1) the longer a predicted category must
be kept in memory before the prediction is satisfied,  the greater the
cost for maintaining that prediction and 2) the greater the distance 
between an incoming word and the most local head or dependent to which 
it attaches,  the greater the integration cost.
Gibson claims his theory explains a wide range of processing complex
phenomena not previously accounted for by a single theory. 
Lee and Wilks (1999) \cite{LW} suggest that it is implausible that there
is a highly nested belief structure computing the nature of speech acts,
rather there is a minimal set of beliefs.
The garden path model \S5.4 uses a minimum principle.
\subsection{Minimal Principles and Word Order.}
If a principle and parameters approach is going to be used in order to
legislate movement between the F-level and the sentence level,  then some 
order must be given to both the elements of the formal declarants and the 
formal string.   It is difficult to justify such an order {\it a priori}:  an 
optimistic hope is that it could be explained by a minimum encoding of 
information and thus to economy of processing.
Sentences,  both formal and informal,  can contain redundant
information.   For example in the predicate calculus strings which are always
true can be added to a given string without effecting the resulting truth 
value.  It is hard to see how this could be compatible with a minimum 
encoding of information.   Also sentences can be expressed in several ways.  
For example in the predicate calculus using the four connectives 
\{and, or, not, implies\},   
or by using the one connective \{Sheffer stroke\}  
or \{Pierce symbol\},
see for example Prior (1962) \cite{bi:prior} p.31.
For the purposes of the P-model in \S\ref{sec:pmodel} it is assumed that an 
ordering of the familiar (or standard) type as used in \S\ref{sec:intro}
can be used.
For other approaches to word order see Downing and Noonan (1995) \cite{bi:DN},
and Bozsahin (1998) \cite{bi:bozsahin}.
\subsection{The Maximal Encoding of Information.}
It is suggested that a maximal amount 
of information about a language's grammar and lexicon (vocabulary) are stored.
This is also an extremal principle however it is the precise opposite of 
more common minimal principles.   
From the point of view of language acquisition,  see \S1.2,
what happens is that when a {\em lexical item} (a particular sample word)
is first heard understanding of it is limited,  so that it is only 
partially learnt,  resulting in it being only used in limited contexts.
As exposure to the lexical item increases more about its semantic and
grammatical properties are learnt so that it can be used in wider contexts.
This fits in with aptitude tests for word meaning,  
where understanding of word nuance is
more important than understanding esoteric words.
What is happening here is that a {\em maximum encoding} of information about
the word is taking place.   Generalizing,  this mechanism happens 
not only to lexical items,  but to many other aspects of language,  
such as understanding of intricate grammatical structures,
this is how linguistic {\it performance} is learnt,  
compare Garman (1990) \cite{bi:garman} \S3.1.2.
At first sight requiring maximal information about a lexical item
might seem inefficient and contrary to economy principles,  
but could occur because it allows 
for a quick analysis of corrupted data,  such as
speech,  which is often only partial heard.
Also maximal storage would aid the very fast processing of language.
What is a minimum and what is a maximum has to be kept track of:
although the information stored about a lexical item might be maximal,
the method of obtaining this information could be minimal.
\section{Psycholinguistics Models of Word Production.}
\subsection{Serial verses Connectionist models of word production.}
Psychological models of a segment (or part) of a language come in two 
basic types:  serial and connectionist.   Serial models work like a serial 
computer programme with each operation being performed sequentially.  
Connectionist models have objects which interact 
to alter one another's connection weights.   
These ideas can be applied to whole sentences or individual words.   
For the T-model or something similar to work there must be serial processing 
at a late stage in sentence production,  because of the discrete 
nature of its lexical items and representations,  see diagram one.
This does not preclude connectionist processing 
before the D-structure representation,  or even before 
the understanding of an individual word.   
There are PROLOG models of restricted sentence production,
Johnson and Klein (1986) \cite{bi:JK1,bi:JK2}.
\subsection{The Stemberger Interactive model of individual word production.}
\label{stemsubsec}
There are psycholinguistic models of individual word production.
There is evidence that the reception of speech is interactive,  
for example there is evidence that seeing the speaker speak influences the 
word heard,  McGurk and McDonald (1976) \cite{bi:McGMcD}; also 
Tanenhaus {\it et al} (1995) \cite{TSES} examine how visual context 
influences spoken word recognition and mediated syntactic processing,  
even during the earliest moments of language processing.
Furthermore there is evidence that verbal
speech production interacts with gesture,  McNeil (1985) \cite{bi:mcneil},  
and various other physiological activities,
Jacobson (1932) \cite{bi:jacobson} p.692,  and these observations 
suggests that the verbal part of speech production is interactive.  A model 
which allows both phonological and semantic influences to interact is the 
psycholinguistic interaction speech production model,  c.f. Stemberger 
(1985) \cite{bi:stemberger}.   
In this model,  when the word ``feather'' is activated a lot of other 
words are also activated with varying weights according to how closely they 
resemble ``feather''.   This can be pictured by diagram four,  to quote   
Stemberger's p.148 text:  
\begin{figure}
   \centering
   {\epsfig{figure=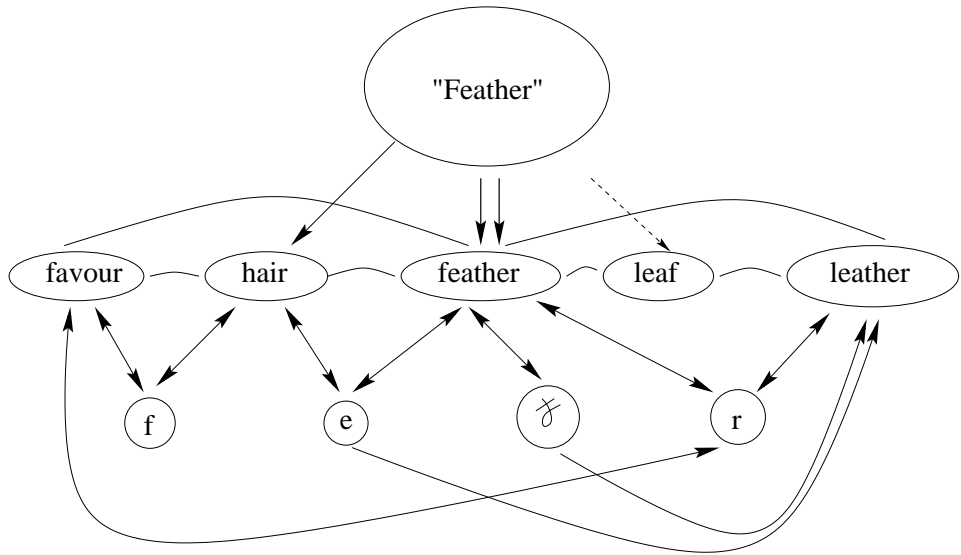,width=1.0\textwidth}}
   \caption{Diagram Four - The Stemberger Diagram.}
\end{figure}
\begin{quote}
Semantic and phonological effects on lexical access.
... an arrow denotes an activating link,  while a filled circle is an 
inhibitory link.   A double line represents a large amount of activation,  
a single solid line somewhat less,  and a broken line even less.   Some of the
inhibitory links have been left out ... for clarity.   The exact nature of 
semantic representation is irrelevant here,  beyond the assumption that it is 
composed of features;   ... a word in quotation marks represents its meaning.
\end{quote}
There is suppression (also called inhibition) across a level,  and 
activation up or down to the next level.   This model accounts for syntax 
by giving different weights to the different words so that words on the left 
come first.   Speech errors come from the noise in the system.   There are 
{\it three} kinds of noise.   
The {\it first} is that the resting level of a unit node 
is subject to random fluctuations;  with the result that it is not the case 
that the unit nodes degree of activation remains at the base line level.
A fluctuation could produce a random production of a part of a word.   
The {\it second} is that words that are used with a high frequency 
have a higher resting level,  and therefore reach activation threshold,  
or "pop out",  quicker.   
This implies that there should be less error for these high frequency words;  
furthermore it implies that when real words occur as an error,  higher 
frequency words should occur as errors more often,  and this does not happen.  
The {\it third} is the so-called systematic spread of activation;   
this means that  the weights in the interaction allow an inappropriate 
activation of word.   There are connectionist programmes of word recognition,
e.g. Seidelberg and McClelland (1989) \cite{bi:SM}.
\subsection{Atomist Semantic Feature Models.}
There are atomist semantic feature models 
(Garman (1990) \cite{bi:garman} p.388) to
which serial models of individual word meaning could be built,  however the 
connectionist interaction picture seems to have less drawbacks.   From now on 
it is assumed that words in some concrete form can be assumed and the question
becomes how are they related to make longer structure such as sentences.
\subsection{The Marslen-Wilson model of word recognition.}
Marslen-Wilson (1987) \cite{marslen}
claims that the process of spoken word recognition breaks down into 
three basic functions:  access,  selection and integration.
{\em Access} is concerned with the mapping of the speech input 
onto the representations of lexical form.
At the earliest stages of the lexical access process,  all lexical memory
elements whose corresponding words begin with a particular acoustic sequence
will be fully activated,   these activated words are called a {\sc cohort},
see also Garman (1990) \cite{bi:garman} p.286.
{\em Selection} is concerned with the discrimination of the best fitting
match to this input.
{\em Integration} is concerned with the mapping of syntactic and semantic
information at the lexical level onto higher levels of processing.
Context comes into play after there is information about the syntactic 
and semantic properties of members of the cohort.
Such models embody the concepts of multiple access and multiple assessment,
allowing for a maximally efficient recognition process,
based on the principle of the contingency of perceptual choice.
\subsection{The classification approach to word prediction.}
Zohar and Roth (2000) \cite{ZR} say that
the eventual goal of a language model is to accurately predict 
the value of a missing word given its context. 
They present an approach to word prediction that is based on learning 
a representation for each word as a function of words and linguistics 
predicates in its context.
They address a few questions that this approach raises.
{\it Firstly} in order to learn good word representations 
it is necessary to use an expressive representation of the context. 
They present a way that uses external knowledge to generate expressive 
context representations,   along with a learning method capable of handling 
the large number of features generated this way that can, potentially, 
contribute to each prediction. 
{\it Secondly} since the number of words ``competing'' for each prediction 
is large, there is a need to ``focus the attention'' on a smaller
subset of these. 
They exhibit the contribution of a ``focus of attention'' mechanism 
to the performance of the word predictor. 
Finally they describe a large scale experimental study in which 
the approach presented is shown to yield significant improvements 
in word prediction tasks. 
\section{Psycholinguistic Models of Sentence Production.}
\subsection{Sentence production in larger structure.}
Research in psycholinguistics can be split up according to the size of the
structure under study,  for example text processing,  sentence processing 
and word meaning.   
Here structure larger than sentences,  for example discourse 
(see Graesser {\it et al} \cite{GMZ}),  are not looked at.
Sentence processing is concerned with how the syntactic 
structures of sentences are computed,  and text processing is concerned 
with how the meanings of larger units of text are understood.
Research in both domains has begun to use the information that can be 
obtained from a large corpora of naturally occurring texts.
In text processing,  recent research has focused on what information the 
words and ideas of a text evoke from long term memory quickly,  passively,  
and at low processing cost;  text processing is not looked at here.
According to McKoon and Ratcliff (1998) \cite{MR}
in recent sentencing research,  a new and controversial theme is that syntactic
computations might rely heavily on statistical information about the relative 
frequencies with which different syntactic structures occur in language.
Gerdemann and van Noord (1999) \cite{GN} discuss various rewrite rules used 
in several areas of natural language processing,
such rewrite rules might change word order.
Hall (1995) \cite{hall} discusses the representations of various linguistic
competences.
Sentences can mean different things in different contexts,
see Akman and Surav (1998) \cite{AS} 
and MacDonald {\it et al} (1994) \cite{MPS}.
Ferro {\it et al} (1999) \cite{FVY} produce learning transformation rules
that find grammatical relations and find that grammatical relations 
between core syntax groups bypasses much of the parsing phrase.
Tabor {\it et al} (1997) \cite{TJT} describe a model which works by analogy
with dynamical systems.   Attractors are taken simultaneously to have 
properties of syntactic categories,  with some encoding of context 
dependent lexical information.   Various experiments were contrived 
which examined the interactions of simple lexical frequencies,  
and their results favoured their dynamical model over traditional approaches.
Truswell {\it et al} (1994) \cite{TTG} devise two eye-movement experiments 
which show that animate nouns where harder to disambiguate when parsing.
\subsection{The Clarke \& Clarke serial model\\ of sentence production.}
There are numerous serial models of sentence production and textbooks
on the subject for example Rosenberg (1977) \cite{bi:rosenberg}.   
Here {\bf three},  the Clarke and Clarke serial model,  the Garret serial
model,  and the garden path model  are very briefly presented 
before going on to the P-model.   
Clarke and Clarke \cite{bi:CC} p.278 mention the 
formulation of an articulatory [sic] program [sic] which has five steps:
\newline
(1){\it Meaning Selection}:   The first step is to decide on the meaning the 
   present constituent is to have.\newline
(2){\it Selection of a Syntactic Outline}:  
The next step is to build a syntactic outline of the constituents.   
It specifies a succession of word slots and indicates which slots 
are to get primary,  secondary,  and zero stress.\newline
(3){\it Content word selection}:   The third step is to select nouns,  verbs,  
   adjectives,  and adverbs to fit the appropriate slots.\newline
(4){\it Affix and function word formation}:   
With the content words decided on, 
the next step is to spell out the phonological shape of the function words 
 (like articles,  conjunctions,  and prepositions),  prefixes,  and suffixes.
\newline
(5){\it Specification of phonetic segments}: 
The final step is to build up fully
specified phonetic segments syllable by syllable.
\newline
By {\it step (5)},  the articulatory program [sic] is complete 
and can be executed.
Typically,  however,  people monitor what they actually say to make certain 
it agrees with what they intended it to mean.   Whenever they detect an error,
they stop,  correct themselves,  and then go on.   It seems likely that the 
more attention is required elsewhere - in planning of various sorts - the less
likely they are to detect an error.   Indeed,  many tongue-slips go unnoticed 
by both speakers and listeners.
\subsection{The Garret serial model of sentence production.}
The Garret (1980) \cite{bi:garret} 
(see also Garman (1990) \cite{bi:garman} p.394) has three basic levels:  
the message level,  the sentence level,  and the articulatory level.   At the 
message level a mental model or image of what is about to be expressed is 
formulated.   Loose ideas of the form of the individual words and the overall 
structure in which they are expressed are formulated to give the sentence
level.   Here the actual words to be used and the structure in which they 
occur are crystallized to give the positional level representation which is 
articulated.
\subsection{The Garden Path Model.}
More recent garden path models are reviewed in Frazier (1987) \cite{frazier},
who says on pages 561-562 (slightly adjusted):\\
``In the garden path model,  perceivers incorporate each word of an input into
a constitute structure representation of the sentence,  
roughly as each item is encountered.   At each step in this process,  the
perceiver postulates the minimal number of nodes required by the grammar of the
language under analysis,  given the structure assigned to preceding items.
This leads to the {\em two} principles of the garden path model:\\
(1){\it Minimal Attachment}:  Do not postulate unnecessary nodes.\\
(2){\it Late Closure}:  If grammatically permissible,  attach new items 
into the clause or phrase currently being processed 
(i.e. the phrase or clause postulated most recently).''\\
In other words minimal attachment entails that a perceiver, 
given so much of the beginning of a sentence
chooses the minimal completion of it which makes sense both grammatically 
and semantically.   The minimal attachment principle is related to the 
requirement that the ultrametric height of a sentence should be a minimum,
see Roberts (1998) \cite{ultra} \S3,  
which was written before I had heard of the garden path model.   
If sentences have \={X} structure and hence binary branching at each node,
see for example Roberts (1998) \cite{ultra} \S2, 
then the two notions are the same.
Minimal attachment is also in accord with the 
minimal principles discussed in \S3.1.
McRae {\it et al} (1998) \cite{MS} use time measuring experiments to see how
event specific knowledge resolves structural ambiguity.   Their results
suggest that the structure of sentences is best described by 
a semantic constraint model,  
then a garden path model with a very short delay,  
and finally by a one region delay garden path model.
Their models and experiments show that event specific knowledge 
is used immediately in sentence comprehension,  and this agrees
with maximal encoding of information in \S3.3 above.
\section{The P-model.}
\label{sec:pmodel}
\subsection{Description of the P-model.}
The P-model can be pictured by the diagram five.
\begin{figure}
   \centering
   {\epsfig{figure=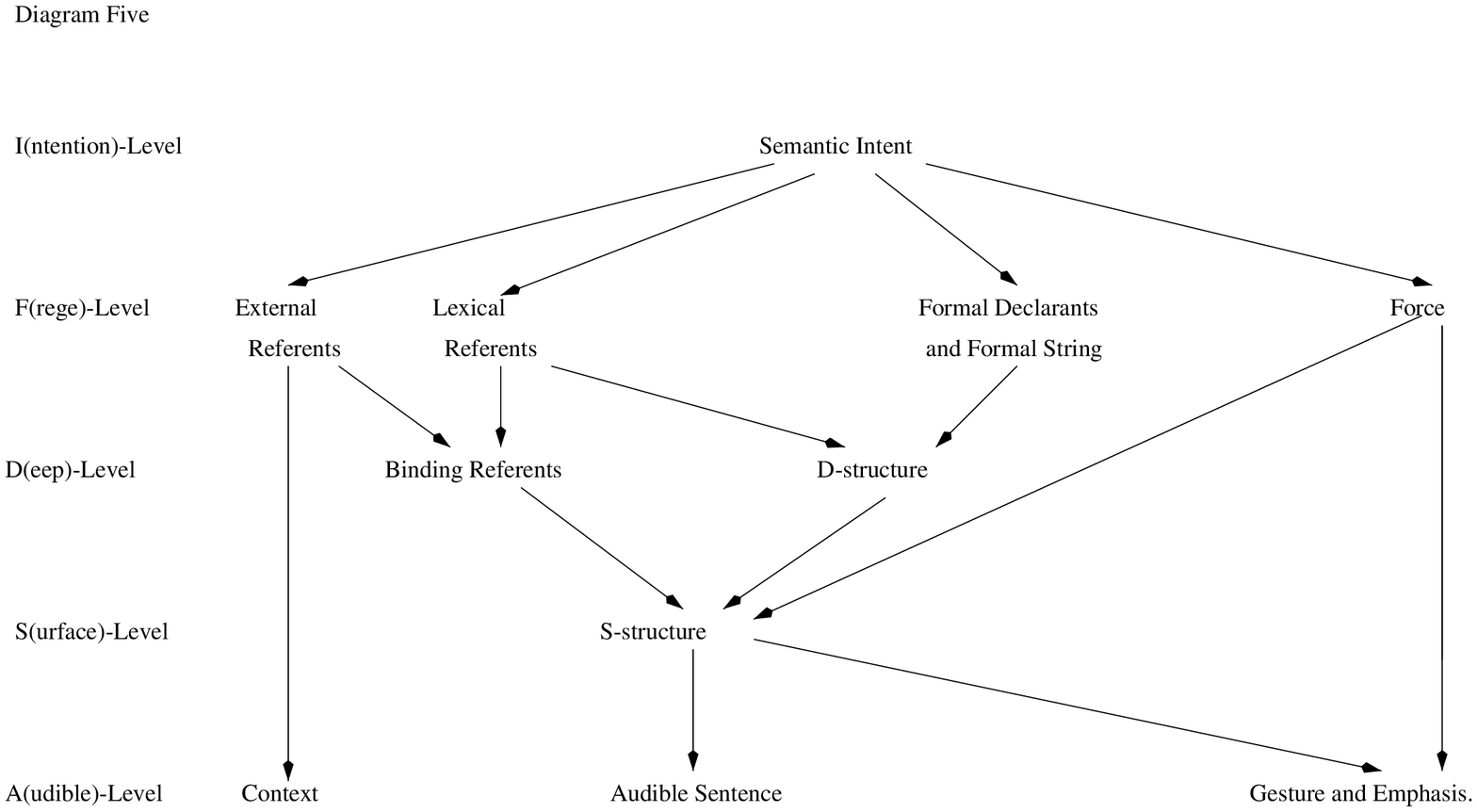,width=1.0\textwidth,height=0.7\textheight}}
   \caption{Diagram Five - The P-model.}
\end{figure}
Semantic intent produces a F-level representation of the sentence and its 
context and force.   
Thought (semantic intent),  occurs prior to words (lexical referents).   
The external referents and the lexical referents combine
to produce binding referents which constrain movement between the D-structure
and S-structure representation.   
The D-structure is constructed from the lexical
referents (words) and the formal declarants and string.   The reason that the 
D-structure sentence changes to the S-structure sentence is because of the
emphasis (or Fregean Force) the speaker wishes to convey in the sentence.
It is known,  see the beginning of \S\ref{stemsubsec},
that in some cases gesture and other behaviour co-occur with spoken 
sentences:  hence the Fregean force and the S-structured sentence interact 
to produce the gesture and emphasis concomitant with the audible production
of the sentence.   It is possible to produce variants of the above analysis.
\subsection{The P-model from a principle \& parameters viewpoint.}
From a principles and parameters view point:  for the T-model movement
between S-structure and LF has to be explained,  
however for the P-model movement
between the F-level and D-structure has to be explained.   Disregarding the 
caveats of \S2 concerning LF,  the P-model requires an explanation of
movement between LF and D-structure,  as depicted by diagram six.   
This can be illustrated by choosing similar examples to \S1.4,1.5,
but now suggesting quantifier lowering and Wh-lowering.
\subsection{Quantifier Lowering.}
This can be represented by diagram six
\begin{figure}
   \centering
   {\epsfig{figure=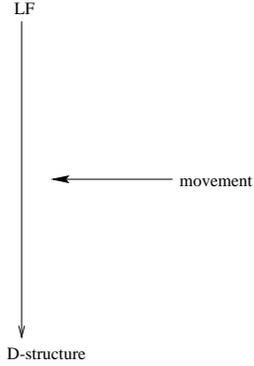,height=0.5\textwidth}}
   \caption{Diagram Six - Quantifier Lowering}
\end{figure}
or in terms of symbols
\be
\forall x(H\varepsilon x \rightarrow J \varepsilon Sx)
\ee
where H,  J and S denote ``human'',  ``Jones'',  and ``saw'' as in \S1.4.
This can be expressed in words
\be
y_{i}~ Jones~ saw~ everyone_{i}
\ee
or in terms of the quantifier $everyone_{i}$
\be
Everyone~ Jones~ saw.
\ee
where $y_{i}$ the trace left by the movement of the word $everyone_{i}$.
\subsection{Wh-lowering.}
This can again be represented by diagram six,  and corresponds to \S1.5.  
In term of symbols
\be
Wh~(x),x\varepsilon H,J\varepsilon Sx.
\ee
which can be expressed in words
\be
y^{i}~ did~ Jones~ see~ who^{i}?
\ee
or filling in the trace
\be
Who~ did~ Jones~ see?
\ee
\section{Conclusion.}
The T-model is explicitly a serial sentence production model.
It is the basic framework in which modern linguistics is currently understood,
linguists being mainly occupied with describing movement between D-structure 
and S-structure.   In psychology there are also serial 
sentence production models which bear little resemblance to the T-model.   
Rather than try to directly accommodate the T-model into a psycholinguistic 
serial model here an entirely new model has been constructed 
based upon altering the T-model from what the author perceives to be its 
defects.   One of the defects is that the T-model requires that a S-structure 
sentence can,  after it has been uttered,  be analyzed into a formal component 
called LF.   This analysis requires movement analogous to that between 
D-structure and S-structure.   
There is no apparent sentence production purpose for doing 
such a {\it post-hoc} analysis,  
the reason seems to be that historically given sentences were 
analyzed for their logical content.   Here it was argued that the correct 
place for LF is before D-structure.   This fits in naturally into a processing
model of sentence production,  where the LF can be thought of as part of the 
message level.  Despite expressing reservations about string order in LF 
(at least as it is commonly understood),  it was shown in some simple cases 
how movement can be described between LF and D-structure.   In this case 
description of movement has a purpose as it elucidates part of a sentence 
production model.   From the point of view of a linguist the most important
requirement of any model is whether there is movement between a 
D-structure and S-structure. This is retained in the P-model.

\end{document}